\documentclass[11pt,draftcls,onecolumn,journal]{IEEEtran}
\usepackage{amsmath,amsfonts}
\usepackage{array}
\usepackage[font=footnotesize]{caption}
\usepackage{multirow}
\usepackage{tikz,calc}
\usetikzlibrary{shapes,arrows,positioning,calc}
\usepackage[caption=false,font=normalsize,labelfont=sf,textfont=sf]{subfig}
\usepackage{textcomp}
\usepackage{stfloats}
\usepackage{url}
\usepackage{verbatim}
\usepackage{graphicx}
\usepackage{cite}
\usepackage{hyperref}
\usepackage{booktabs}
\usepackage{xcolor}
\usepackage[ruled,vlined]{algorithm2e}
\usepackage[framemethod=tikz]{mdframed}
\input{mysymbol.sty}
\tikzstyle{phantom vertex} = [ ellipse, 
                               anchor = center, 
                               minimum height = 0.06*\unit, 
                               minimum width  = 0.06*\unit,
                               inner sep=0pt,
                               anchor=center]
\tikzstyle{red vertex}   = [black, fill = red!20,   phantom vertex, draw]
\tikzstyle{black vertex} = [black, fill = black!20, phantom vertex, draw]
\tikzstyle{blue vertex}  = [black, fill = blue!20,  phantom vertex, draw]
\tikzstyle{green vertex} = [black, fill = green!20,  phantom vertex, draw]
\tikzstyle{vertex}       = [draw, phantom vertex]

\tikzstyle{point} = [ellipse, inner sep=0pt, draw, fill=white, anchor = center,
                     minimum height = 0.05*\unit, minimum width  = 0.05*\unit]
\begin{document}

\title{Concomitant DAG Learning: 
On the Roles of Noise Adaptivity, Sparsity, and Non-negativity 
}

\author{Gonzalo Mateos,~\IEEEmembership{Senior~Member,~IEEE}, 
Samuel Rey,~\IEEEmembership{Member,~IEEE},
Hamed Ajorlou,~\IEEEmembership{Graduate Student~Member,~IEEE}, and Mariano Tepper$^{\dagger}$
\thanks{$^{\dagger}$G. Mateos and H. Ajorlou are with the Dept. of Electrical and Computer Engineering at the
University of Rochester, Rochester, NY (emails: \href{gmateosb@ece.rochester.edu}{gmateosb@ece.rochester.edu}, \href{hajorlou@ur.rochester.edu}{hajorlou@ur.rochester.edu}). S. Rey is with the Dept. of Signal Theory and Communications,
Rey Juan Carlos University, Madrid, Spain (e-mail: \href{samuel.rey.escudero@urjc.es}{samuel.rey.escudero@urjc.es}). M. Tepper is with Elastic, San Francisco, CA (email:\href{mariano.tepper@gmail.com}{marianotepper@gmail.com}). Manuscript received May 22, 2026; revised \today.}
}

\markboth{IEEE Signal Processing Magazine,~Vol.~XX, No.~XX, October~2026 (REVISED)}%
{Mateos \MakeLowercase{\textit{et al.}}: Concomitant DAG Learning}

\maketitle

\vspace{-2.5cm}




\section*{Introduction}
\label{sec:scope}
Directed acyclic graphs (DAGs) have well-appreciated merits for encoding causal relationships, with a long history in the context of Bayesian networks~\cite[Ch. 1.2.2]{Pearl_2009}. This graphical modeling paradigm is central to signal processing (SP) and machine learning (ML) applications spanning domains such as biology~\cite{sachs}, healthcare~\cite{bio}, genetics~\cite{genetics}, and finance~\cite{finance}, 
to name a few. However, since the causal structure underlying a group of variables is often unknown, there is a need to address the task of inferring DAGs from (passive) observational data. While additional interventional data are provably beneficial to the related problem of causal discovery~\cite{intervention4,elements}, said interventions oftentimes are infeasible or ethically challenging to implement~\cite{DSL2022survey}; see also~\cite{peng2025isit} for recent advances involving soft interventions. Learning a DAG solely from observational data poses significant computational challenges, primarily due to the combinatorial acyclicity constraint which is notoriously difficult to enforce~\cite{np2}. Moreover, distinguishing between DAGs that generate the same observational data distribution is nontrivial. This fundamental identifiability challenge may arise when data are limited (especially in high-dimensional settings), or, when candidate graphs in the search space belong to the same equivalence class ({i.e.}, they exhibit the so-termed Markov equivalence). In addition, several conditions ({e.g.,} sufficiency, faithfulness, minimality) must be satisfied so that causal interpretations can be made from the sought DAG; see e.g.,~\cite[Ch. 7.1]{elements}.

Since DAG learning from observational data is in general an NP-hard problem, recent efforts have advocated a continuous relaxation framework which offers an efficient means of exploring the space of DAGs~\cite{noTears,golem,dagma,saboksayr2023colide}, albeit without global optimality guarantees. In addition to breakthroughs in smooth characterizations of acyclicity, an appropriate score function (composed of data fitting and regularization terms) is paramount to guide gradient-based algorithms towards good-quality solutions. 
Workhorse ordinary least squares (OLS) implicitly relies on homoscedasticity~\cite{loh2014high}, meaning the variances of the exogenous noises in postulated structural equation models (SEMs) are identical across variables. 

The pursuit of parsimonious DAG representations dictates that score functions often include a \emph{sparsity}-promoting regularization, for instance an $\ell_1$-norm penalty as in lasso regression. This in turn necessitates careful fine-tuning of the penalty parameter that governs the trade-off between sparsity and data fidelity \cite{generalized-con-lasso}. Theoretical insights suggest a regularization weight proportional to the observation noise level \cite{lasso2}, but the latter is typically unknown. Accordingly, several linear regression studies (unrelated to DAG learning) have proposed convex \emph{concomitant} estimators based on scaled LS, which jointly estimate the noise level along with the regression coefficients \cite{belloni2011sqrtlasso,con-lasso1,smooth-con-lasso}. A noteworthy representative is the smoothed concomitant lasso \cite{smooth-con-lasso,generalized-con-lasso}, which not only addresses the aforementioned parameter fine-tuning predicament but also accommodates heteroscedastic scenarios. While these challenges are not extraneous to DAG learning, the impact of concomitant estimators is so far largely unexplored in this timely field that is woven into the natural fabric of the process that entails ``going from signals to causes''.

{In this context, this tutorial aims to delineate the analytical background and SP relevance of \emph{structure-aware} statistical inference and optimization tools to tackle the combinatorial problem of learning DAGs from observational data. Distinct forms of structure we will explore include edge sparsity, non-negative edge weights, and noise heteroscedasticity.} 
We start with SEMs as multivariate causal models. Those will be used to formally state the DAG structure learning problem, and to selectively survey key results and assumptions for structure identifiability (i.e., uniqueness). We will then offer a brief historical overview of DAG structure identification methods, with an emphasis on so-termed \emph{score-based} approaches for linear SEMs. 
Our didactic exposition will introduce various loss functions used to score candidate DAGs, along with innovative exact acyclicity characterizations. 
With these foundations in place, we bring to bear ideas from concomitant scale estimation in penalized linear regression and introduce a novel convex score function for sparsity-aware DAG inference~\cite{saboksayr2023colide}. A key message is that by making the score function \emph{noise adaptive}, one effectively removes the coupling between the sparsity regularization parameter and the unknown exogenous noise levels, which leads to minimum (or no) recalibration effort across diverse problem instances or distribution shifts. Relative to OLS, 
we demonstrate that CoLiDE (\textbf{Co}ncomitant \textbf{Li}near \textbf{D}AG \textbf{E}stimation) is better suited for challenging heteroscedastic scenarios. Finally, we discuss the impact of additional DAG structure on top of sparsity; namely, \emph{non-negativity} of edge weights~\cite{rey2026nonnegativity}. {In applications where this assumption is tenable (e.g., when DAGs are presumed to be binary or devoid of inhibitory effects), non-negativity constraints have favorable optimization landscape and statistical implications, which we also bring to light along with algorithmic opportunities.} We conclude with a summarizing discussion as well as an outline of CoLiDE extensions and other frontier topics that offer exciting research outlooks at the confluence of (graph) SP, ML, optimization, and causal inference. {The core material surveyed in this article comes from~\cite{saboksayr2023colide,rey2026nonnegativity}, which we reframe to offer new insights.}




\section*{Preliminaries and DAG structure learning problem statement}
\label{sec:prob_statement}

We start with graph-theoretic properties of DAGs and then reveal their key role in encoding conditional independencies among observable variables in directed graphical models (Bayesian networks), whose joint distribution satisfies a Markov property with respect to (w.r.t.) the DAG. We introduce SEMs to generate said Markovian distributions, and state the inverse problem of learning multivariate causal models, namely recovering the DAG structure from samples of the joint distribution alone. For concreteness, our focus throughout this tutorial will be (mostly) on linear SEMs, and we close this section by reviewing key results and assumptions guaranteeing structure identifiability for this class of problem instances. 

\subsection*{DAGs, SEMs, and topology inference problem formulation}
\label{ssec:dags_sems_prob}

Consider a directed (di)graph $\ccalG \left ( \ccalV, \ccalE, \bbW \right)$, where $\ccalV=\{1,\ldots,d\}$ represents the set of vertices, and $\ccalE \subseteq \ccalV \times \ccalV$ is the edge set. The {(weighted)} adjacency matrix $\bbW = [\bbw_1, \ldots, \bbw_d] \in \reals^{d \times d}$ collects the edge weights, with $W_{ij} \neq 0$ indicating a direct link from node $i$ to node $j$. We henceforth assume that $\ccalG$ belongs to the space $\mathbb{D}$ of DAGs, and rely on the graph to represent conditional independencies among entries in the random vector $\bbx = [ x_1, \ldots, x_d ]^{\top}\in \reals^d$. If the joint distribution $p(\bbx)$ satisfies a Markov property w.r.t. $\ccalG\in\mathbb{D}$, each {nodal} variable $x_i$ depends solely on its parents $\textrm{PA}_i := \{j \in \ccalV : W_{ji} \neq 0\}=\textrm{support}(\bbw_i)$. 

\begin{mdframed}[hidealllines=true,backgroundcolor=gray!20]
{\bf Partially-ordered sets, DAGs, and graph signal processing (GSP)}\\ 
\begin{minipage}[t]{0.75\textwidth}
A partially-ordered set (poset) $\ccalV$ with elements $i, j, k \in \ccalV$ satisfies: (i) $i \leq i$ (reflexivity); (ii) $i \leq j$ and $j \leq i$ implies $i = j$ (antisymmetry); and (iii) $i \leq j$ and $j \leq k$ implies $i \leq k$ (transitivity). DAGs and posets are intimately related: every DAG induces a unique partial order on its nodes $\ccalV$, where $j < i$ whenever there is a directed path from $j$ to $i$. Conversely, any poset can be realized by at least one DAG whose edges encode the partial ordering 
\end{minipage}
\begin{minipage}[t]{0.10\textwidth}
\vspace{-0.8cm}
\begin{center}
\def \myfactor{0.7}
{\fontsize{8}{8}\selectfont
\begin{tikzpicture}[auto, node distance=1.25cm, every loop/.style={},
                    thick,main node/.style={circle,draw}, scale = \myfactor]
\tikzstyle{stuff_fill}=[node,draw,fill=black!20]

\node[main node] (1) [label=above:\red{$x_{1}$},draw=blue!80,fill=yellow!30] {1};
\node[main node] (2) [below left of=1] [label=left:\red{$x_{2}$},draw=blue!80,fill=yellow!30] {2};
\node[main node] (3) [below right of=1] [label=right:\red{$x_{3}$},draw=blue!80,fill=yellow!30] {3};
\node[main node] (4) [below right of=2] [label=left:\red{$x_{4}$},draw=blue!80,fill=yellow!30] {4};
\node[main node] (5) [below of=4] [label=below:\red{$x_{5}$},draw=blue!80,fill=yellow!30] {5};

\draw [->, color=blue] (1) -- node [swap] {\green{$W_{12}$}} (2);
\draw [->, color=blue] (1) -- node {\green{$W_{13}$}} (3);
\draw [->, color=blue] (1) -- node {\green{$W_{14}$}} (4);
\draw [->, color=blue] (1) to [out=30,in=90] ($(3.east)+(1.25cm, -0.5cm)$) to [out=-90,in=-15] node {\green{$W_{15}$}}(5);
\draw [->, color=blue] (2) -- node [swap] {\green{$W_{24}$}} (4);
\draw [->, color=blue] (3) -- node {\green{$W_{34}$}} (4);
\draw [->, color=blue] (4) -- node [swap] {\green{$W_{45}$}} (5);
\draw [->, color=blue] (3) to [out=-70,in=30] node {\green{$W_{35}$}} (5);


\end{tikzpicture}}
\end{center}
\end{minipage}

\vspace{11pt}
\noindent relation. In a poset, not all elements are necessarily comparable. For the DAG in the figure,  e.g., node {$1 < 5$} while nodes {$2$ and $3$} cannot be ordered. In the depicted example the nodes happen to be topologically sorted, and hence the adjacency matrix $\bbW$ will be strictly upper triangular. 

Since DAGs introduce a partial order over $\ccalV$, it is prudent to ask how to incorporate this inductive bias into information processing architectures for $\bbx$ {(a DAG signal in the GSP parlance~\cite{mateos2019connecting})}. GSP tools like the graph Fourier transform are not directly applicable, since the adjacency matrix of a DAG has all its eigenvalues equal to zero and, hence, lacks an eigenbasis. An elegant framework that extends GSP to DAGs and posets was put forth in~\cite{seifert2023causal}; see also~\cite{rey2026dcn}.  
\end{mdframed}

Here, we focus on \emph{linear} SEMs to generate the so-termed observational distribution $p(\bbx)$, whereby the relationship between each random variable and its parents in the DAG is given by
\begin{equation}\label{eq:linear_sem_nodal}
x_i = \bbw_i^{\top} \bbx + z_i,\quad \:i\in\ccalV,
\end{equation}
where $\bbz = [z_1, \ldots, z_d]^{\top}$ is a vector of mutually independent, exogenous noises; see e.g.,~\cite{elements}.  Noise variables $z_i$ can have different variances $\sigma_i^2$ and need not be Gaussian distributed. As we shall soon elucidate, (non)linearity and noise distributional assumptions are central to model identifiability. Identity \eqref{eq:linear_sem_nodal} is analogous to a discrete-time Markov chain, where the next state only depends on some function of the current state and i.i.d. noise. For a dataset $\bbX\in \reals^{d \times n}$ of $n$ i.i.d. samples drawn from $p(\bbx)$, the linear SEM assignments \eqref{eq:linear_sem_nodal} can be written in compact matrix form as $\bbX = \bbW^{\top}\bbX + \bbZ$. Beyond i.i.d. measurements, we also encounter linear structural vector autoregressive models (SVARMs) for time series data in the methods presented in the ``Non-negative edge weights'' section. We will briefly touch upon modeling of nonlinear mechanisms as one of the frontier topics we discuss in our concluding remarks.\vspace{2pt}

\noindent{\bf Problem statement.} Given the data matrix $\bbX\in \reals^{d \times n}$ adhering to a linear SEM, our goal is to learn the latent DAG $\ccalG\in \mathbb{D}$ described by its adjacency matrix $\bbW$ as the solution to the optimization problem
\begin{equation}\label{eq:prblem_hard}
    \min_{\ccalG(\bbW)} \;\; \ccalS (\ccalG(\bbW);\bbX) \;\; \text{subject to} \;\; \ccalG(\bbW) \in \mathbb{D},
\end{equation}
where $\ccalS$ is a data-dependent score function to measure the quality of candidate graphs. Irrespective of the criterion, the non-convexity comes from the combinatorial acyclicity constraint $\ccalG\in \mathbb{D}$.

A proper score function typically encompasses a loss or data-fidelity term ensuring alignment with the SEM as well as regularizers to promote desired properties of $\ccalG$. For a linear SEM, the OLS loss $\tfrac{1}{2n} \| \bbX - \bbW^{\top} \bbX \|_F^2$ is widely adopted, where $\|\cdot\|_{F}$ is the Frobenius norm~\cite{loh2014high}. The choice $\tfrac{1}{2n} \| \bbX - \bbW^{\top} \bbX \|_1$ was also advocated under the premise of sparse root causes, motivated by the signal model in~\cite{seifert2023causal}. When $\bbZ$ are Gaussian with unknown diagonal covariance, the data log-likelihood can be an effective loss~\cite{golem}. Sparsity is a cardinal property of most graphs, so one can penalize the OLS loss with an $\ell_1$-norm as in
\begin{equation}\label{eq:ols_l1_loss}
\ccalS (\ccalG(\bbW);\bbX) = \tfrac{1}{2n}\| \bbX - \bbW^{\top} \bbX \|_{F}^2 + \lambda \| \bbW \|_1,
\end{equation}
where $\lambda\geq 0$ is a tuning parameter that controls edge sparsity. The score \eqref{eq:ols_l1_loss} resembles the \emph{multi-task} variant of lasso regression, specifically when the response and design matrices coincide. Optimal minimax rates for lasso hinge on selecting $\lambda \asymp \sigma \sqrt{\log d / n}$ \cite{lasso2}. However, the exogenous noise variance $\sigma^2$ is rarely known in practice. This challenge is compounded in heteroscedastic settings, where one should adopt a \emph{weighted} LS score (see~\cite{loh2014high} for an exception unlike most DAG learning methods that stick to bias-inducing OLS). Recognizing these limitations, in the ``Concomitant DAG estimation'' section we survey a noise-adaptive LS-based score function for joint estimation of the DAG and the noise levels.

\subsection*{Structure identifiability}
\label{ssec:identifiability}

In their most general form, SEMs consist of a collection of $d$ structural assignments
\begin{equation}\label{eq:general_sem_nodal}
x_i = f_i(\bbx_{\textrm{PA}_i},z_i),\quad \:i\in\ccalV,
\end{equation}
where $\bbx_{\textrm{PA}_i}$ denotes the entries of $\bbx$ indexed by node $i$'s parents $\textrm{PA}_i\subset \ccalV$ in the DAG, and the noise variables $z_i$ are mutually independent. This model class is overly expressive, in the sense that given a joint distribution $p(\bbx)=\Pi_{i=1}^dp(x_i\given \bbx_{\textrm{PA}_i})$ that satisfies the Markov property w.r.t.  $\ccalG\in\mathbb{D}$, there always exists an SEM \eqref{eq:general_sem_nodal} that entails $p(\bbx)$~\cite[Prop. 7.1]{elements}. Apparently, one needs additional restrictions on the model to uniquely recover the DAG structure $\bbW$ from the observational distribution $p(\bbx)$. 
In the remainder of this section, we review some selected assumptions and conditions that lead to structure identifiability. Some of these results can be quite technical and we only offer informal statements conveying the key conceptual ingredients; for full details the interested reader is referred to the cited references. 

A Markovian distribution $p(\bbx)$ is \emph{faithful} w.r.t. the underlying DAG $\ccalG\in\mathbb{D}$ when there is a bijection between conditional independencies in the distribution and Pearl's graph-related ``$d$-separation'' statements~\cite[Def. 1.2.3]{Pearl_2009}. A typical manifestation of non-identifiability arises when multiple DAGs encode the same conditional independencies. Collectively, these graphs comprise a Markov equivalence class (MEC) and they are represented by a completed partially DAG (CPDAG){~\cite[Definition 6.24]{elements}. In CPDAGs, an edge will be directed only when all MEC members contain the edge in that direction. Otherwise, the CPDAG edge is undirected.} Under faithfulness, the MEC (i.e., the CPDAG, but not $\ccalG$) is identifiable from $p(\bbx)$~\cite[Lemma 7.2]{elements}, and this is the principle behind \emph{constraint-based} methods that test conditional independencies from data.

Most germane to our exposition in this tutorial, consider narrowing the general model \eqref{eq:general_sem_nodal} to the class of linear SEMs introduced in \eqref{eq:linear_sem_nodal}. As it turns out, this restriction alone is still insufficient to ensure uniqueness. Non-identifiable instances arise when $\bbz$ has a multivariate Gaussian distribution (so-termed linear Gaussian SEMs), except in the homoscedastic setting where the noise entries $z_i$ have equal noise variances; see e.g.,~\cite[Prop. 7.1]{elements} and~\cite{loh2014high}. For linear SEMs with non-Gaussian distributed errors, dubbed linear non-Gaussian acyclic models (LiNGAMs), identifiability has been established using independent component analysis (ICA)~\cite[Th. 7.6]{elements}; see also~\cite{loh2014high} for an alternative proof when noise variances are known up to a common multiplicative factor. Nonlinearity can also be fruitfully exploited, since Gaussian (additive) SEMs $x_i = f_i(\bbx_{\textrm{PA}_i})+z_i$, with smooth nonlinear mechanisms $f_i$ are identifiable~\cite[Th. 7.7]{elements}.

With these foundations in place, we switch gears to statistical methods and associated (optimization) algorithms developed to estimate $\ccalG\in\mathbb{D}$ from finite data $\bbX$ via the score-minimization formulation \eqref{eq:prblem_hard}.




\section*{Structure Identification Methods: A Historical Overview}
\label{sec:structure_identification_review}

We review the evolution of \emph{score-based} DAG learning methodology (constraint-based methods will also be briefly outlined for completeness), charting the progression from classic combinatorial search to modern continuous optimization frameworks. Early score-based methods deal with a formidable search over superexponentially (in the number of variables) many DAGs, typically using parametric likelihood-based scores and greedy heuristics to navigate $\mathbb{D}$ with affordable complexity but limited guarantees of global optimality beyond small toy problems. A transformative paradigm shift occurred with the advent of continuous relaxations that recast DAG learning as constrained optimization over real-valued adjacency matrices endowed with smooth characterizations of acyclicity; see~\cite{DSL2022survey} for a recent tutorial treatment surveying structure identification approaches for \emph{nonlinear SEMs} as well. 

\begin{mdframed}[hidealllines=true,backgroundcolor=gray!20]
{\bf Constraint-based approaches via conditional independence testing}\\ 
Constraint-based (a.k.a. independence-based) methods rely on the assumption that the distribution $p(\bbx)$ is faithful w.r.t. the DAG, in which case the MEC described by a CPDAG is uniquely identifiable. The one-to-one correspondence between $d$-separation and conditional independence statements entailed by $p(\bbx)$ naturally suggests statistical inference procedures to probe candidate graph (sub)structures. Indeed, most constraint-based methods such as the celebrated PC algorithm~\cite{spirtes2001causation} first conduct conditional independence tests to estimate the undirected skeleton. They then orient as many edges as possible by identifying $v$-structures in the CPDAG {(i.e., triads where two non-adjacent vertices are parents of a common node)~\cite[Lemma 6.25]{elements}}; a complete set of orientation rules is known. Admittedly, testing for conditional independencies from finite data is a non-trivial endeavor in its own right, especially in high-dimensional settings. Under Gaussian assumptions, one can equivalently test for null partial correlations~\cite{mateos2019connecting,giannakis18} and in this setting the PC algorithm can be shown to yield a consistent estimator of CPDAG structure; see~\cite[p. 146]{elements} and references therein.
\end{mdframed}

\subsection*{Discrete optimization methods}
\label{ssec:discrete_methods}

A broad swath of approaches falls under the category of score-based methods, where the basic premise is to assess candidate DAGs in terms of their suitability to describe the observations in $\bbX$. To this end and recalling~\eqref{eq:prblem_hard}, often a parametric model is assumed (say, a linear Gaussian SEM) and a customized score function $\ccalS (\ccalG(\bbW);\bbX)$ is used to guide the search for DAGs in $\mathbb{D}$. Typical choices include complexity-penalized likelihood functions such as the Bayesian information criterion (BIC) or minimum description length (MDL)-based alternatives; Bayesian scoring functions are also popular~\cite[Chapter 7.2.2]{elements}. Irrespective of the score chosen, these \emph{combinatorial optimization} methods exhibit scalability issues and finding the best scoring DAG is typically intractable. This is because the number of possible DAGs $|\mathbb{D}_d|$ grows at a superexponential rate with the number of variables $d$. There are $|\mathbb{D}_3|=25$ different DAGs on $d=3$ variables and this number balloons to $|\mathbb{D}_8|=783702329343$ as per the recursion
\begin{equation*}
|\mathbb{D}_d|=\sum_{i=1}^d(-1)^{i-1}\binom{d}{i}2^{i(d-i)}|\mathbb{D}_{d-i}|, \quad |\mathbb{D}_{0}|:=1.
\end{equation*}
Aiming at tractability, search techniques introduce modifications to the original problem by incorporating additional assumptions on the DAG, such as bounding the number of parent nodes associated with each variable~\cite{disc3}. Greedy search strategies are widely adopted, whereby given a candidate $\ccalG\in\mathbb{D}$ one evaluates the score on a small set of neighboring DAGs (for instance, those obtained by adding, removing, or flipping a single arc in $\ccalG$). A new best-scoring candidate is chosen from these neighbors; otherwise, the procedure terminates if no improvement can be made. For (non-identifiable) linear Gaussian SEMs, the greedy equivalence search (GES) algorithm~\cite{greedy1} explores the space of MECs rather than individual DAGs using a BIC criterion, and can be shown to recover the true CPDAG asymptotically as $n\to\infty$. 

\subsection*{Order-based methods}\label{ssec:order_based_methods}

A related (still challenging) problem is to determine the causal order among the observed variables in $\ccalV$. If the DAG's topological ordering were known, then the structure identification task would boil down to classical variable selection. Indeed, for a given node $i$ we must have $\textrm{PA}_i\subseteq\{1,\ldots,i-1\}$, and the parents can be found using, e.g., sparse regression based on \eqref{eq:linear_sem_nodal} and added constraints on $\bbw_i$. This rationale motivated other recent \emph{order-based} approaches that exploit the neat equivalence $\ccalG(\bbW) \in \mathbb{D} \Leftrightarrow \bbW = \bbPi^{\top} \bbU \bbPi$, where $\bbPi \in \{0,1\}^{d \times d}$ is a binary permutation matrix (essentially encoding the topological ordering) and $\bbU \in \reals^{d \times d}$ is an upper-triangular matrix of edge weights. Consequently, one can search over exact DAGs by formulating an end-to-end differentiable minimization of $\ccalS (\bbPi^{\top} \bbU \bbPi;\bbX)$ jointly over $\bbPi$ and $\bbU$, or in two steps. Even for nonlinear SEMs, determining the appropriate node ordering has been attempted through the Birkhoff polytope of permutation matrices, using techniques like the Gumbel-Sinkhorn approximation~\cite{two2}, or, the SoftSort operator~\cite{two1}. Limitations stemming from misaligned forward and backward passes that respectively rely on hard and soft permutations are well-documented~\cite{permutahedron}. The DAGuerreotype approach in~\cite{permutahedron} instead searches over the permutahedron of \emph{vector} orderings, while TOPO~\cite{topo} performs a bilevel optimization that relies on topological order swaps at the outer level. Despite significant progress, optimization challenges towards accurately recovering the causal ordering remain, especially when data are limited and the noise level profile is heterogeneous.

\subsection*{Continuous optimization}
\label{ssec:cont_relax}

A flurry of recent methods advocate an exact algebraic characterization of acyclicity using smooth nonconvex functions $\ccalH:\reals^{d\times d}\mapsto \reals$ of the adjacency matrix $\bbW$, whose zero level set is $\mathbb{D}$. The elegant idea is then to relax the combinatorial constraint $\ccalG(\bbW) \in \mathbb{D}$ by instead enforcing $\ccalH(\bbW)=0$, 
\begin{equation}\label{eq:discrete_cont}
\min_{\ccalG(\bbW)} \;\; \ccalS (\ccalG(\bbW);\bbX)\;\;
		\text{subject to} \;\; \ccalG(\bbW) \in \mathbb{D}
		\quad \iff \quad \min_{\bbW} \;\; \ccalS (\bbW;\bbX)\;\;
		\text{subject to} \;\; \ccalH(\bbW)=0
\end{equation}
and tackle the DAG learning problem \eqref{eq:prblem_hard} using standard gradient-based optimization algorithms that scale to larger graphs~\cite{noTears, poly, noFears, dagma}. Notice that we slightly abused the notation in \eqref{eq:discrete_cont} as the leftmost score function takes on a graph argument, while $\ccalS(\bbW;\bbX)$ is defined on the space of $d\times d$ matrices. 

\begin{figure}
\begin{minipage}{0.70\textwidth}
		\begin{equation*} 
        e^{\bbW} = \sum_{k=0}^{\infty} \frac{\bbW^k}{k!}  = \begin{bmatrix}
			1 & 0 & 0 \\
			0 & 1 & 0 \\
			0 & 0 & 1
		\end{bmatrix} + \underbrace{\begin{bmatrix}
				0 & 1 & 0 \\
				0 & 0 & 1 \\
				1 & 0 & 0
		\end{bmatrix}}_{\text{edges/self-loops}} +  \frac{1}{2}\underbrace{\begin{bmatrix}
				0 & 0 & 1 \\
				1 & 0 & 0 \\
				0 & 1 & 0 
		\end{bmatrix}}_{\text{length-2 paths/cycles}} +  \frac{1}{6}\underbrace{\begin{bmatrix}
				\blue{1} & 0 & 0 \\
				0 & \blue{1} & 0 \\
				0 & 0 & \blue{1} \end{bmatrix}}_{\text{length-3 paths/cycles}}+ \cdots 
         \end{equation*}       
\end{minipage}
\hfill
\begin{minipage}{0.13\textwidth}
\centering
\def \myfactor{0.7}
{\fontsize{8}{8}\selectfont
\begin{tikzpicture}[auto, node distance=1.25cm, every loop/.style={},
                    thick,main node/.style={circle,draw}, scale = \myfactor]
\tikzstyle{stuff_fill}=[node,draw,fill=black!20]

\node[main node] (1) [label=above:$x_{1}$,draw=blue!80,fill=yellow!30] {1};
\node[main node] (2) [below left of=1] [label=above:$x_{2}$,draw=blue!80,fill=yellow!30] {2};
\node[main node] (3) [below right of=1] [label=above:$x_{3}$,draw=blue!80,fill=yellow!30] {3};

\draw [->, color=blue] (1) -- (2);
\draw [->, color=blue] (2) -- (3);
\draw [->, color=blue] (3) -- (1);

\end{tikzpicture}}
\end{minipage}
\caption{Illustrating the NOTEARS acyclicity characterization with a toy binary graph $\ccalG$. The matrix exponential is a polynomial series in the adjacency matrix $\bbW$, and the $i$th diagonal entry of $\bbW^k$ counts the number of length-$k$ cycles in $\ccalG$ that include node $i$. The depicted graph has a cycle of length $3$ reflected in the non-zero diagonal of $\bbW^3$, so $\ccalH_{\text{expm}}(\bbW)=1/2$. The term $\bbW^0=\bbI$ adds a spurious count of $d$ for all graphs, hence the correction in $\ccalH_{\text{expm}}(\bbW)$ which vanishes if and only if $\ccalG$ is acyclic.}
\label{fig:matrix_exponential}
\end{figure}

The pioneering NOTEARS formulation~\cite{noTears} proposed
\begin{equation}\label{eq:notears}
\ccalH_{\text{expm}}(\bbW) = \operatorname{tr}(e^{\bbW \circ \bbW}) - d,
\end{equation}
where $\circ$ denotes the Hadamard (element-wise) product and $\operatorname{tr}(\cdot)$ is the matrix trace operator. Diagonal entries of powers of $\bbW \circ \bbW$ encode information about cycles in $\ccalG$, hence the suitability of the chosen function; see also Figure \ref{fig:matrix_exponential} {for a binary DAG example}. The Hadamard product squares all edge weights while preserving the support of $\bbW$, thus avoiding missed cycles due to (rare but possible) catastrophic cancellation of sign-indefinite weights over closed paths. The closed-form gradient $\nabla \ccalH_{\text{expm}}(\bbW) = 2\bbW\circ (e^{\bbW \circ \bbW})^\top$ can be evaluated in $O(d^3)$ complexity using highly-optimized scientific computing libraries developed to operate with matrix exponentials. Follow-up work suggested a more computationally efficient acyclicity function $\ccalH_{\text{poly}}(\bbW) = \operatorname{tr}((\bbI + \frac{1}{d} \bbW \circ \bbW)^{d}) - d$, where $\bbI$ is the identity matrix~\cite{poly}. Motivated by the Cayley-Hamilton theorem, the general family $\ccalH(\bbW) = \sum_{k=1}^d c_k\operatorname{tr}((\bbW \circ \bbW)^{k})$, $c_k> 0$, was studied in~\cite{noFears}, which subsumes $\ccalH_{\text{poly}}(\bbW)$ as a special case when $c_k=\binom{d}{k}/d^k$. 

Recently, the nilpotency property of DAGs inspired DAGMA's log-determinant regularizer~\cite{dagma}
\begin{equation}\label{eq:dagma}
\ccalH_{\text{ldet}}(\bbW; s) = d \log(s) - \log(\det(s \bbI - \bbW \circ \bbW)),\:\: s>0.
\end{equation}
The function $\ccalH_{\text{ldet}}(\bbW; s)$ is an exact acyclicity characterization over the domain $\mathbb{W}^s=\{\bbW\in\reals^{d\times d}:s>\rho(\bbW\circ\bbW)\}$, where $\rho(\cdot)$ denotes the spectral radius of a square matrix. Because adjacency matrices of DAGs have all-zero eigenvalues, $\mathbb{D}\subset\mathbb{W}^s$ for any $s>0$. The gradient is given by $\nabla \ccalH_{\text{ldet}}(\bbW; s) = 2\bbW\circ(s\bbI - \bbW \circ \bbW)^{-\top}$. DAGMA outperformed prior continuous relaxation methods in terms of (nonlinear) DAG recovery and computational efficiency due to several factors, among which we highlight improved gradient behavior and the ability to detect longer cycles; see~\cite[Section 3.2]{dagma} for further details. {While NOTEARS~\cite{noTears} relies on augmented Lagrangian methods for constrained optimization, DAGMA instead advocates a barrier-type solver~\cite[Section 4]{dagma} we later review.} {In the ``Non-negative edge weights'' section, we argue that the added structural assumption of non-negativity can bring around important simplifications (one can drop the Hadamard product) to DAGMA's acyclicity function $\ccalH_{\text{ldet}}$ along with further optimization benefits, which suggest reconsidering method of multipliers (MoM) solvers for \eqref{eq:discrete_cont}.}

\subsection*{Performance evaluation metrics}
\label{ssec:metrics}

We close by listing standard metrics used for evaluating the performance of DAG learning algorithms. While we focus on assessing edge set identification (i.e., the support of $\bbW$), accurate estimation of edge weights is important and typically evaluated using the score functions described so far. For all metrics other than the true positive rate (TPR) and the F1-score, a \emph{lower} value $(\downarrow)$ indicates better performance.

\noindent {\bf True positive rate (TPR).} Proportion of correctly identified edges relative to the total number of edges in the ground-truth DAG.

\noindent {\bf False discovery rate (FDR).} Ratio of incorrectly identified edges to the total number of detected edges.

\noindent {\bf F1-score.} Harmonic mean between precision and recall (a.k.a. TPR), where precision is the proportion of correctly identified edges relative to the total number of detected edges.

\noindent {\bf Structural Hamming distance (SHD).} {Total count of edge additions, deletions, and reversals required to map the estimated graph into the ground-truth DAG. Often normalized by the number of nodes $d$.} 

\noindent {\bf SHD on CPDAGs (SHD-C).} Adopted to evaluate recovery performance of the MEC. Initially, we map both the estimated graph and the ground-truth DAG to their corresponding CPDAG. 
Subsequently, we calculate the SHD between the two CPDAGs to yield the SHD-C metric. 
    
\noindent{\bf Structural intervention distance (SID).} Quantifies the number of disrupted causal paths in the predicted DAG by counting (over vertex pairs) the number of falsely inferred intervention distributions{; see~\cite{peters2015sid}}. 




\section*{Concomitant DAG estimation}
\label{sec:colide}

Going back to our discussion about lasso-type score functions for SEMs, minimizing \eqref{eq:ols_l1_loss} subject to a smooth acyclicity constraint $\ccalH(\bbW)=0$ as in, e.g., NOTEARS~\cite{noTears} or DAGMA~\cite{dagma}: (i) requires laborious retuning of $\lambda$ when the unknown $\sigma^2$ changes across problem instances; and (ii) implicitly relies on limiting homoscedasticity assumptions due to the OLS criterion. To address issues (i)-(ii), here we review a continuous optimization approach that for the first time integrates concomitant estimation of scale parameters into a causal structure identification pipeline~\cite{saboksayr2023colide}. Specifically, CoLiDE introduces a convex score function $\ccalS(\bbW,\sigma;\bbX)$ to \emph{jointly} estimate the DAG adjacency matrix and exogenous noise standard deviation(s). The upshot is a \emph{noise-adaptive} procedure that is robust (both in terms of DAG learning performance and parameter fine-tuning) to possibly heteroscedastic exogenous noise profiles.
The methodology was inspired by the literature of concomitant sparse linear regression \cite{con-lasso1,smooth-con-lasso,generalized-con-lasso}, {dating back to Huber's seminal work on robust joint location and scale estimators that we also highlight~\cite{robust1,con-lasso1}.} The idea is simple, elegant, and of independent interest -- yet in our view not widely known across SP circles. We will also touch upon optimization considerations for this challenging problem. 

\subsection*{Homoscedastic setting}
\label{ssec:colide_ev}

Consider the homoscedastic scenario in which all exogenous noises $z_1,\ldots,z_d$ in the linear SEM \eqref{eq:linear_sem_nodal} have identical variance $\sigma^2$. Building on the smoothed concomitant lasso~\cite{smooth-con-lasso}, the problem of jointly estimating the DAG adjacency matrix $\bbW$ and the exogenous noise scale $\sigma$ can be formulated as~\cite{saboksayr2023colide}
\begin{equation}\label{eq:genral-colide-ev}
    \min_{\bbW, \sigma\geq \sigma_0} \; \underbrace{\frac{1}{2 n \sigma} \| \bbX - \bbW^{\top}\bbX \|_{F}^2 + \frac{d \sigma}{2} + \lambda \| \bbW \|_1}_{:=\ccalS(\bbW,\sigma;\bbX)} \;\;
    \text{subject to } \;\; \ccalH(\bbW)=0,
\end{equation}
where $\ccalH:\reals^{d\times d}\mapsto \reals$ is a smooth nonconvex function, whose zero level set is $\mathbb{D}$ as discussed in the ``Continuous optimization'' section. Notably, the weighted, regularized LS score function $\ccalS(\bbW,\sigma;\bbX)$ in \eqref{eq:genral-colide-ev} is now also a function of $\sigma$, and it can be traced back to Huber's robust linear regression work~\cite{robust1} that we review in ``Concomitant estimation of scale: A robust statistics redux''. {A few remarks are in order. First, notice that the squared residuals are normalized by $\sigma$, not $\sigma^2$ as in weighted LS. Due to this specific rescaling [cf. \eqref{eq:ols_l1_loss}], $\lambda$ in \eqref{eq:genral-colide-ev} decouples from $\sigma$ as minimax optimality now requires $\lambda \asymp \sqrt{\log d / n}$ \cite{li2020fast}.} {Second, while $\ccalS(\bbW,\sigma;\bbX)$ is jointly convex in $\bbW$ and $\sigma$, \eqref{eq:genral-colide-ev} is still a nonconvex optimization problem due to the acyclicity constraint $\ccalH(\bbW)=0$. Third, Huber reverse-engineered the additive term $d\sigma/2$ so that the resulting variance estimator is consistent under Gaussianity; see \eqref{eq:sigma-ev-hat}. Lastly,} the constraint $\sigma \geq \sigma_0$ safeguards against potential ill-posed scenarios where the estimate $\hat{\sigma}$ approaches zero{; see ~\cite[Sec. 2.3]{smooth-con-lasso} for a technical justification.} {Following the guidelines in~\cite{smooth-con-lasso}, one can set $\sigma_0 = \frac{ \| \bbX \|_F}{\sqrt{dn}} \times 10^{-2}$.}

Regarding the choice of the acyclicity function, the formulation in~\cite{saboksayr2023colide} opts for the DAGMA penalty $\ccalH_{\text{ldet}}$ in \eqref{eq:dagma} based on several compelling reasons we stated in the ``Continuous optimization'' section. Moreover, while LS-based linear DAG learning approaches {such as} NOTEARS are prone to introducing long spurious cycles in the estimated graph, it was noted that a log-determinant term arising with the Gaussian log-likelihood score used in GOLEM~\cite{golem} tends to mitigate this undesirable effect. Interestingly, the same holds true when $\ccalH_{\text{ldet}}$ is chosen as a regularizer, but this time without being tied to Gaussian assumptions. Before moving on to optimization issues, we emphasize that CoLiDE is in principle flexible to accommodate other convex loss functions beyond LS (e.g., Huber's loss $h_M$ as in \eqref{eq:concomitant_huber} for robustness against heavy-tailed contamination), other acyclicity functions, and even nonlinear SEMs parameterized using e.g., neural networks. We elaborate on these and several other extensions in the ``Discussion and the road ahead'' section that concludes the article.

\begin{mdframed}[hidealllines=true,backgroundcolor=gray!20]
{\bf Concomitant estimation of scale: A robust statistics redux}\\ 
\begin{minipage}[t]{0.8\textwidth} Consider a linear regression setting with measurements $y_i=\bbbeta^\top\bbx_i+\varepsilon_i$, for $i=1,\ldots,n$. OLS is well motivated for Gaussian errors $\varepsilon_i$ with common variance $\sigma^2$, but in the presence of outliers Huber suggested the celebrated loss $h_{M}(z)$  
\end{minipage}
\begin{minipage}[t]{0.10\textwidth}
\vspace{-1cm}
\begin{center}
{\fontsize{8}{8}\selectfont
			\begin{tikzpicture}[domain=-2:2, scale=0.7]
				\draw[->] 		 (-2.2, 0.0) -- ( 2.2,0.0) node[right] {$z$};
				\draw[->] 		 ( 0.0,-0.3) -- ( 0.0,2.5) node[at end, right] {$h_M(z)$} node[at start, right] {$0$};
				
				\draw[gray, dashed] 		 (1, 1) -- (1,0)  node[at end, below, black] {$M$};
				
				\draw[gray, dashed] 		 (-1, 1) -- (-1,0) node[at end, below, black] {$-M$};
				
				\draw [red, domain=-1:1] plot (\x, {(\x)*(\x))});

				\draw [blue, domain=1:1.5] plot (\x, {2*(\x)-1)});
				
				\draw [blue, domain=-1.5:-1] plot (\x, {-2*(\x)-1)});
		\end{tikzpicture}}
\end{center}
\end{minipage}

\vspace{11pt}
\noindent
in the figure, which downweighs the effects of large residuals exceeding $M\sigma$ in magnitude. Parameter $M$ is typically fixed to e.g., $M=1.35$ for maximum robustness, while guaranteeing $95\%$ efficiency under Gaussianity. So in practice one has to (robustly) estimate the scale parameter $\sigma$ along with the regression coefficients $\bbbeta$. To this end, Huber proposed the concomitant estimator~\cite{robust1}
\begin{equation}\label{eq:concomitant_huber}
\min_{\bbbeta,\sigma}\left[ \sum_{i=1}^n h_M\left(\frac{y_i-\bbbeta^\top\bbx_i}{\sigma}\right)\sigma+n\sigma\right],
\end{equation}
which for fixed $\sigma$ yields the usual robust coefficients $\bbbeta$ under the loss $h_M$. {Interestingly, \eqref{eq:concomitant_huber} is jointly convex in $\{\bbbeta,\sigma\}$ by virtue of \cite[Lemma 5.2]{con-lasso1} and the convexity of affine compositions.}


\noindent An observation worth sharing was made in~\cite{con-lasso1} with regards to Huber's concomitant estimator for the quadratic loss. Replacing $h_M(z)$ with $\rho(z)=z^2$ in \eqref{eq:concomitant_huber}, one finds the minimizers are given by the OLS coefficients $\hbbeta$ and $\smash{\hat{\sigma}^2=\frac{1}{n}\sum_{i=1}^n (y_i-\hbbeta^\top \bbx_i)^2}$ -- the maximum likelihood estimates for Gaussian errors. This is remarkable, as Huber effectively ``convexified'' the negative log likelihood 
\begin{equation*}
\ccalL(\bbbeta,\sigma)=\frac{n}{2}\log 2\pi +n\log \sigma +\frac{1}{2\sigma^2}\sum_{i=1}^n (y_i-\bbbeta^\top\bbx_i)^2
\end{equation*}
which is not convex in $\sigma$, hence it cannot be jointly convex in $\{\bbbeta,\sigma\}$.

{\noindent Over the last two decades, concomitant lasso estimators also gained traction for sparse regression. Originally proposed in~\cite{con-lasso1}, they were rediscovered under various denominations such as the square-root lasso~\cite{belloni2011sqrtlasso}, scaled lasso~\cite{con-lasso2}, and smoothed concomitant lasso~\cite{smooth-con-lasso}. Beyond practical convenience in the selection of sparsity regularization parameters, noise adaptivity via joint scale estimation is of interest for uncertainty quantification and model selection using e.g., BIC scores that depend on $\sigma$.}
\end{mdframed}

\subsection*{Optimization considerations}
\label{ssec:colide_opt}

Motivated by the choice of the acyclicity function, the constrained optimization problem \eqref{eq:genral-colide-ev} can be tackled following the DAGMA methodology in~\cite{dagma}. Therein, it is suggested to solve a sequence of \emph{unconstrained} problems where $\ccalH_{\text{ldet}}$ is viewed as a regularizer. This technique has proven more effective in practice when compared to, say, an augmented Lagrangian method used in NOTEARS~\cite{noTears}. Given a decreasing sequence $\mu_k\to 0$, at step $k$ of the COLIDE-EV (equal variance) algorithm one solves
\begin{equation}\label{eq:colide-ev}
    \min_{\bbW, \sigma \geq \sigma_0} \; \mu_k \left[ \frac{1}{2 n \sigma} \| \bbX - \bbW^{\top}\bbX \|_{F}^2 + \frac{d \sigma}{2} + \lambda \| \bbW \|_1 \right] + \ccalH_{\text{ldet}}(\bbW, s_k),
\end{equation}
where a schedule of hyperparameters $\mu_k\geq 0$ and $s_k>0$ must be prescribed prior to implementation. Decreasing $\mu_k$ enhances the influence of the acyclicity function $\ccalH_{\text{ldet}}(\bbW, s)$ in the objective. The sequential procedure \eqref{eq:colide-ev} is reminiscent of the central path approach of barrier methods, and the limit of the central path $\mu_k\to 0$ is guaranteed to yield a DAG~\cite{dagma}. In theory, this means no additional post-processing (e.g., edge trimming) is needed. {However, in practice thresholding helps reduce false positives as we do not get to $\mu_k=0$ in the schedule. The interested practitioner is referred to~\cite[Sec. 4.1]{saboksayr2023colide} for a user-friendly guide on algorithm initialization, hyperparameter selection, and post-processing; see also Algorithm~\ref{A:colide}.}

Unlike DAGMA,  CoLiDE-EV jointly estimates the noise level $\sigma$ and the adjacency matrix $\bbW$ for each $\mu_k$. To this end, one could solve for $\sigma$ in closed form [cf. \eqref{eq:sigma-ev-hat}] and plug back the solution in \eqref{eq:colide-ev}, to obtain a DAG-regularized square-root lasso~\cite{belloni2011sqrtlasso} type of problem in $\bbW$. It is unwise to follow this path because the resulting loss $\|\bbX - \bbW^{\top}\bbX\|_F$ fails to decompose across samples, challenging mini-batch based stochastic optimization if it were needed for scalability, or, envisioned online algorithms for streaming data{; see Figure \ref{fig:App_batch} and its accompanying discussion.} 
A similar issue arises with GOLEM \cite{golem}, where the Gaussian profile likelihood yields a non-separable log-sum loss. Alternatively, (inexact) block coordinate descent (BCD) iterations are advocated in~\cite{saboksayr2023colide}. This cyclic strategy involves fixing $\sigma$ to its most up-to-date value and minimizing \eqref{eq:colide-ev} inexactly w.r.t. $\bbW$, then updating $\sigma$ given the latest $\bbW$ via
\begin{equation}\label{eq:sigma-ev-hat}
    \hat{\sigma} = \max\left(\sqrt{\operatorname{tr}\left( (\bbI - \bbW)^{\top} \operatorname{cov}(\bbX) (\bbI - \bbW)\right)/d},\sigma_0\right),
\end{equation}
where $\operatorname{cov}(\bbX) := \frac{1}{n} \bbX \bbX^{\top}$ is the precomputed sample covariance matrix of the observations. The mutually-reinforcing interplay between noise level and DAG estimation should be apparent. 

There are several ways to inexactly solve the $\bbW$ subproblem using lightweight first-order methods. Indeed, the elegant block successive convex approximation (BSCA) algorithm~\cite{yang2020inexactbcd} can be leveraged to exploit the favorable structure of the concomitant objective, and derive provably convergent iterative solvers with tractable subproblems and favorable per-iteration cost~\cite{COLIDE_SAM_24}; further aligning the CoLiDE framework with scalable SP toolkits for high-dimensional structured inference. Because the required line search can be computationally taxing, the simpler heuristic adopted in~\cite{saboksayr2023colide} is to run a single step of the ADAM optimizer to refine $\bbW$. Experiments show that running multiple ADAM iterations yields marginal gains, since one is anyways continuously re-updating $\bbW$ in the BCD loop. This process is repeated until either convergence is attained (we note that the ADAM-based inexact BCD procedure lacks formal convergence guarantees), or, a prescribed maximum iteration count $T_k$ is reached.

\subsection*{Heteroscedastic setting}
\label{ssec:colide_nv}

We shift gears to the more challenging endeavor of learning DAGs in heteroscedastic scenarios, where noise variables have non-equal variances (NV) $\sigma_1^2,\ldots,\sigma_d^2$. Mimicking the concomitant formulation for the EV case discussed earlier, the CoLiDE-NV estimator is given by
\begin{equation}\label{eq:colide-nv}
    \min_{\bbW, \bbSigma \geq \bbSigma_0} \; \mu_k \bigg[ \frac{1}{2n} \operatorname{tr} \left( (\bbX - \bbW^{\top}\bbX)^{\top} \bbSigma^{-1} (\bbX - \bbW^{\top}\bbX) \right) + \frac{1}{2} \operatorname{tr}(\bbSigma) + \lambda \| \bbW \|_1 \bigg] + \ccalH_{\text{ldet}}(\bbW, s_k).
\end{equation}
{Note that $\bbSigma=\operatorname{diag}(\sigma_1,\ldots,\sigma_d)$ is a diagonal matrix of exogenous noise \emph{standard deviations} (hence not a covariance matrix).} Naturally, \eqref{eq:colide-nv} is backward compatible and simplifies to the legacy COLIDE-EV in \eqref{eq:colide-ev} when $\bbSigma=\sigma\bbI$. 
A closed-form solution for $\bbSigma$ given $\bbW$ is also readily obtained, namely
\begin{equation}\label{eq:sigma-nv-hat}
    \hat{\bbSigma} = \max\left(\sqrt{\operatorname{diag}\left( (\bbI - \bbW)^{\top} \operatorname{cov}( \bbX) (\bbI - \bbW) \right)},\bbSigma_0\right).
\end{equation}
A summary of the overall computational procedure for both CoLiDE variants is tabulated under Algorithm~\ref{A:colide}. CoLiDE's per iteration cost is $\ccalO(d^3)$, on par with state-of-the-art DAG learning methods. 

\begin{figure}
    \centering
    \begin{minipage}{0.61\textwidth}
        \begin{algorithm}[H]\label{A:colide}
        \begin{small}
            \textbf{In:} data $\bbX$ and hyperparameters $\lambda$ and $\ccalP = \{(\mu_k, s_k, T_k) \}_{k=1}^K$.\\
            \textbf{Out:} DAG $\bbW$ and the noise estimate $\sigma$ (EV) or $\bbSigma$ (NV).\\
            Compute lower-bounds $\sigma_0$ or $\bbSigma_0$.\\
            Initialize $\bbW=\mathbf{0}$, $\sigma = \sigma_0 \times 10^2$ or $\bbSigma = \bbSigma_0 \times 10^2$.\\
            \SetInd{0.5em}{0.5em}
            \ForEach{$(\mu_k, s_k, T_k) \in \ccalP$}{
            \For{$t=1,\ldots, T_k$}{
                Apply CoLiDE-EV or NV updates using $\mu_k$ and $s_k$.\\
            }
            }
        \end{small}
        \caption{CoLiDE optimization{~\cite{saboksayr2023colide}}}
        \end{algorithm}
    \end{minipage}%
    \hfill
    \begin{minipage}{0.38\textwidth}
        \setlength{\interspacetitleruled}{0pt}%
        \setlength{\algotitleheightrule}{0pt}%
        \begin{algorithm}[H]\label{A:colide_function}
        \begin{small}
            \SetKwProg{Fn}{Function}{:}{}
            \Fn{CoLiDE-EV update}
            {
                Update $\bbW$ with one iteration of a first-order method for \eqref{eq:colide-ev}\\
                Compute $\hat{\sigma}$ using \eqref{eq:sigma-ev-hat}\\
            }
            
            \Fn{CoLiDE-NV update}
            {
                Update $\bbW$ with one iteration of a first-order method for \eqref{eq:colide-nv}\\
                Compute $\hat{\bbSigma}$ using \eqref{eq:sigma-nv-hat}
            }
        \end{small}
        \end{algorithm}
    \end{minipage}
\vspace{-10pt}
\end{figure}

The first summand in \eqref{eq:colide-nv} resembles the consistent weighted LS estimator studied in \cite{loh2014high}, but therein the assumption is that exogenous noise variances are known up to a constant factor. CoLiDE-NV removes this requirement by estimating $\bbW$ and $\bbSigma$, with marginal added complexity over finding the DAG structure alone. Like GOLEM~\cite{golem} and for general linear Gaussian SEMs, as $n\to\infty$ CoLiDE-NV provably yields a DAG that is {`quasi-equivalent'} to the ground-truth graph{; see~\cite[App. C]{saboksayr2023colide}. The precise notion of (quasi)-equivalence among digraphs was introduced in~\cite{ghassami2020characterizing}, and we will not reproduce all technical details here. It suffices to say that two digraphs are quasi equivalent if the set of distributions that they can both generate has a nonzero Lebesgue measure; the interested reader is referred to~\cite[Section 3.1]{golem}. Understanding the behavior in \emph{high-dimensional} settings where both $n,d\to\infty$ and sparsity will likely play a key role, is a challenging open problem worthy of additional exploration; see also~\cite{loh2014high}.}

\begin{table}[t]\footnotesize
    \caption{DAG recovery results for 200-node ER4 graphs under homoscedastic Gaussian noise. {Reproduced from~\cite{saboksayr2023colide}.}}
    \label{tab:linEV}

    \resizebox{\linewidth}{!}{%
    \begin{tabular}{l cccc cccc}
    \toprule
    & \multicolumn{4}{c}{Noise variance $= 1.0$} & \multicolumn{4}{c}{Noise variance $= 5.0$}\\
    \cmidrule(lr){2-5} \cmidrule(lr){6-9}
    & GOLEM & DAGMA & CoLiDE-NV & CoLiDE-EV & GOLEM & DAGMA & CoLiDE-NV & CoLiDE-EV\\
    \hline
    \hline
    SHD & 468.6$\pm$144.0 & 100.1$\pm$41.8 & 111.9$\pm$29 & {\bf 87.3$\pm$33.7} & 336.6$\pm$233.0 & 194.4$\pm$36.2 & 157$\pm$44.2 & {\bf 105.6$\pm$51.5}\\
    SID & 22260$\pm$3951 & 4389$\pm$1204 & 5333$\pm$872 & {\bf 4010$\pm$1169} & 14472$\pm$9203 & 6582$\pm$1227 & 6067$\pm$1088 & {\bf 4444$\pm$1586}\\
    SHD-C & 473.6$\pm$144.8 & 101.2$\pm$41.0 & 113.6$\pm$29.2 & {\bf 88.1$\pm$33.8} & 341.0$\pm$234.9 & 199.9$\pm$36.1 & 161.0$\pm$43.5 & {\bf 107.1$\pm$51.6}\\
    FDR & 0.28$\pm$0.10 & 0.07$\pm$0.03 & 0.08$\pm$0.02 & {\bf 0.06$\pm$0.02} & 0.21$\pm$0.13 & 0.15$\pm$0.02 & 0.12$\pm$0.03 & {\bf 0.08$\pm$0.04}\\
    TPR & 0.66$\pm$0.09 & 0.94$\pm$0.01 & 0.93$\pm$0.01 & {\bf 0.95$\pm$0.01} & 0.76$\pm$0.18 & 0.92$\pm$0.01 & 0.93$\pm$0.01 & {\bf 0.95$\pm$0.01}\\
\bottomrule
\end{tabular}%
}
\vspace{-10pt}
\end{table}%

\subsection*{CoLiDE in action}
\label{ssec:colide_experiments}

Through {a curated selection of} numerical results across diverse test cases and application domains {from~\cite{saboksayr2023colide}}, here we demonstrate CoLiDE's enhanced stability, scalability and robust accuracy relative to state-of-the-art baselines outlined in the previous section of this article. {Our discussion also exposes some of the method's limitations; see also~\cite[Sec. 5 and App. E]{saboksayr2023colide} for a comprehensive experimental evaluation spanning different graph sizes, edge densities, signal-to-noise ratio levels, and multiple ablation studies.}  

We begin by assuming equal noise variances across all nodes. We employ Erd\H{o}s--R\'enyi (ER4) and scale-free (SF4) DAGs with $d=200$ nodes, average degree of $4$, and edge weights drawn uniformly from the range $[-2, -0.5] \cup [0.5, 2]$~\cite{noTears,golem,dagma}. Results in Table~\ref{tab:linEV} {are reproduced from~\cite{saboksayr2023colide} and} reflect two specific scenarios when the noise is Gaussian: (i) $\sigma^2=1$, as in prior studies \cite{noTears, golem, dagma}; and (ii) when the noise level is increased to $5$. These cases show that CoLiDE's advantage over its competitors is not restricted to SHD alone, but equally extends to all other relevant metrics introduced in ``Performance evaluation metrics''. This behavior is accentuated when the noise variance is set to 5, as CoLiDE naturally adapts to different noise regimes without any manual tuning of its hyperparameters. {Despite being overparametrized for homoscedastic problems, CoLiDE-NV performs well;} either being on par with CoLiDE-EV or the second-best alternative in Table~\ref{tab:linEV}. {This is particularly valuable as the noise characteristics are usually unknown in practice, favoring more general and versatile formulations.}

\begin{figure}[t]
    \centering
    \begin{minipage}[c]{0.83\textwidth}
    \includegraphics[width=\textwidth]{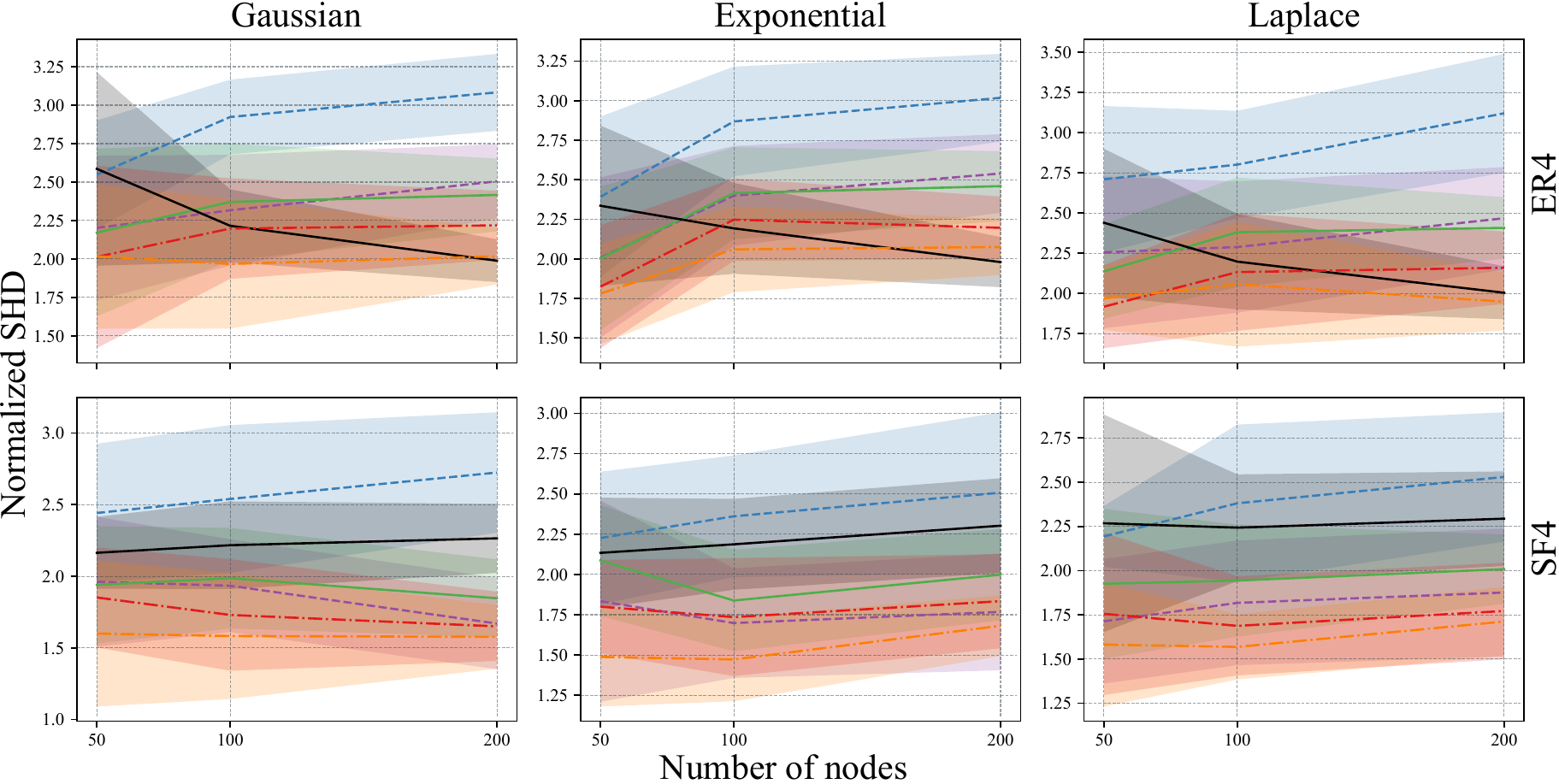}
    \end{minipage}
    \begin{minipage}[c]{0.14\textwidth}
    \includegraphics[width=\textwidth]{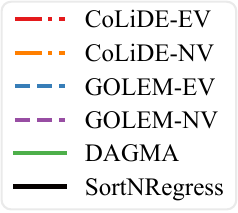}
    \end{minipage}
    \caption{Mean DAG recovery performance, plus/minus one standard deviation, under heteroscedastic noise for both ER4 (top row) and SF4 (bottom row) graphs with varying numbers of nodes. Columns correspond to different noise distributions{~\cite{saboksayr2023colide}}.}
    \label{fig:linNV}
    \vspace{-10pt}
\end{figure}

{\noindent {\bf Heteroscedastic noise.} Further challenges arise when nodes do not share the same noise variance, as e.g., the linear} Gaussian case is known to be non-identifiable~\cite{golem} from observational data. 
This issue is exacerbated as the number $d$ of nodes grows as, whether we are estimating the variances explicitly or implicitly, the problem contains $d$ additional unknowns, which renders its optimization harder from a practical perspective. We select the edge weights by uniformly drawing from $[-1, -0.25] \cup [0.25, 1]$ and the noise variance of each node from $[0.5, 10]$. This compressed interval, compared to the one used in the prior test case, has a reduced signal-to-noise ratio~\cite{noTears,sortnregress}, which obfuscates the optimization.

{Figure~\ref{fig:linNV} reproduces experiments in~\cite{saboksayr2023colide} with varying noise distributions, graph ensembles, and sizes.} CoLiDE-NV outperforms the other baselines in virtually all configurations. {It also consistently maintains lower deviations than DAGMA and GOLEM, underscoring its robustness.} CoLiDE-EV performs well and often is the second-best solution, outmatching GOLEM-NV, even considering that an EV formulation is clearly underspecifying the problem. In this particular instance, SortNRegress~\cite{sortnregress} is competitive with CoLiDE-NV. {Next, we discuss the rationale behind SortNRegress and its impact on benchmarking.}

{\noindent {\bf Data standardization.} In simulated SEMs with additive noise, the data generation model might inadvertently expose the underlying structure, thereby unexpectedly affecting (i.e., implicitly aiding) structure learning algorithms~\cite{sortnregress}. Indeed, the marginal variance of the $x_i$ [cf. \eqref{eq:linear_sem_nodal}] tends to increase along the causal order. This information is exploited in the SortNRegress algorithm proposed in~\cite{sortnregress}, which first sorts nodes based on increasing variance and then identifies parents using lasso regression. To mitigate the information leakage caused by such inadvertent model effects, they also suggest standardizing the data of each node to have zero mean and unit variance. Going back to \eqref{eq:linear_sem_nodal}, one normalizes $x_i$, not the $z_i$.

Such standardization constitutes a nonlinear transformation that can adversely impact performance, particularly for methods developed for linear SEMs. After repeating all experiments following the guidelines in~\cite{sortnregress}, we found that despite the overall degradation in performance resulting from such standardization, CoLiDE-NV continues to outperform state-of-the-art algorithms in the above challenging heteroscedastic task. We also standardized the data from small and sparse ER1 graphs, each comprising $d=20$ nodes, with a substantial number of $n=100,000$ i.i.d. observations. This time, we considered the greedy algorithm GES~\cite{greedy1} for comparisons. Prior to standardization, we obtained the following SHD metrics: CoLiDE-EV: $12.8$; CoLiDE-NV: $14.9$; and GES: $14.9$. Post-standardization, the SHD values changed to CoLiDE-EV: $18.8$; CoLiDE-NV: $19.5$; and GES: $14.9$. Notably, the performance of GES remained unaffected by standardization. While CoLiDE outperformed the other continuous optimization approaches, it displayed some sensitivity to standardization, indicating that it is not as robust in this sense as greedy algorithms. This is a limitation of the CoLiDE approach, and the point raised in~\cite{sortnregress} is well taken.}

\begin{figure}[t]
    \centering
    \begin{minipage}[c]{.45\textwidth}
    \includegraphics[width=\textwidth]{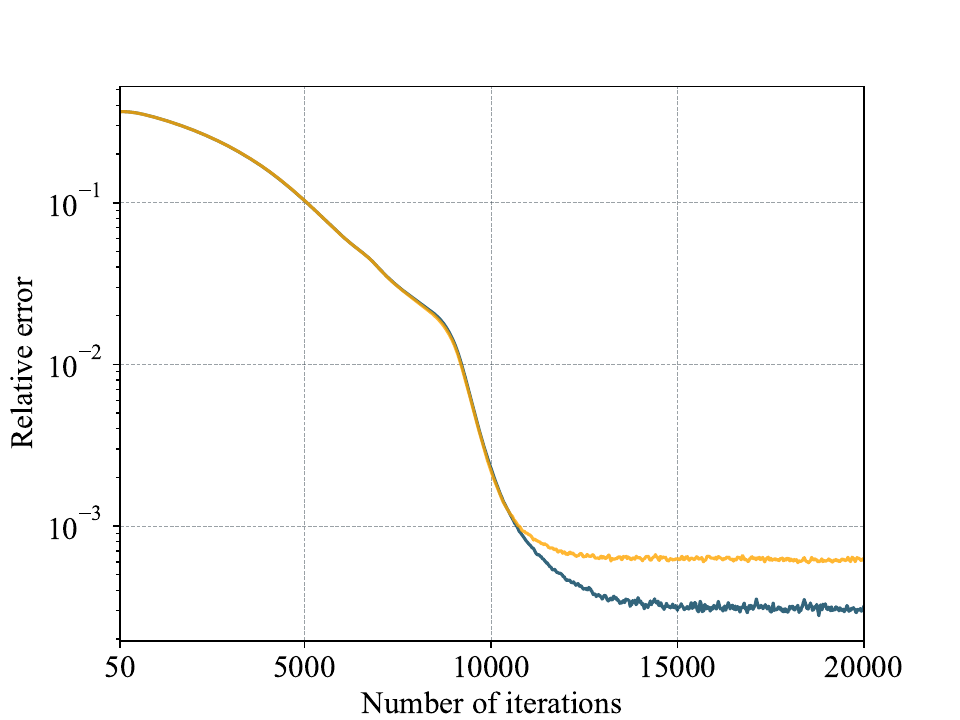}
    \end{minipage}
    \begin{minipage}[c]{.45\textwidth}
    \includegraphics[width=\textwidth]{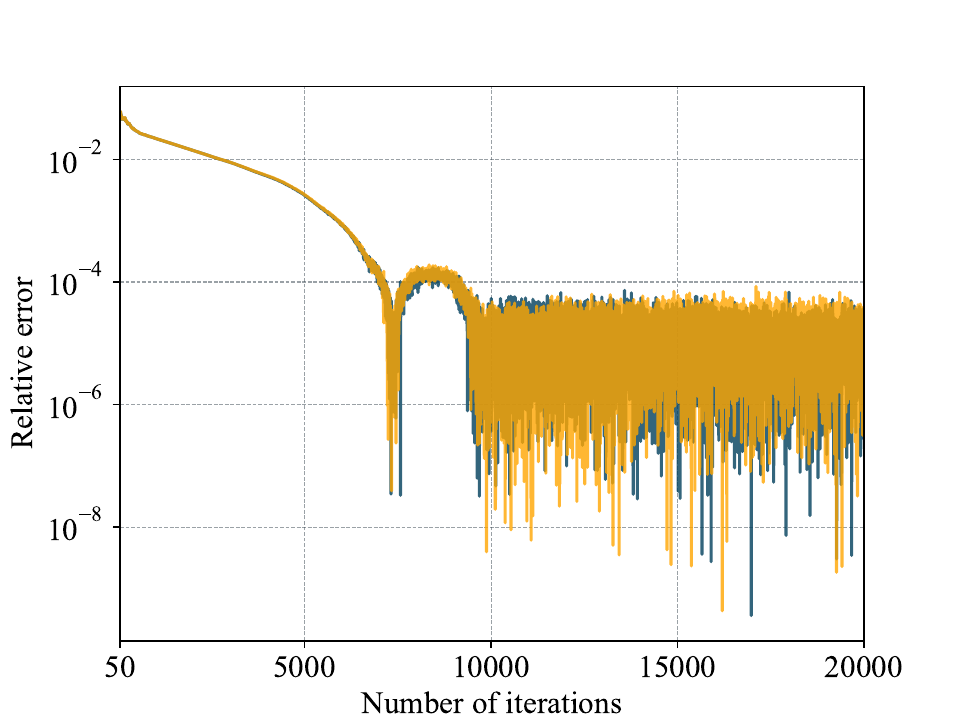}
    \end{minipage}
    \newline
    \begin{minipage}[c]{.55\textwidth}
    \includegraphics[width=\textwidth]{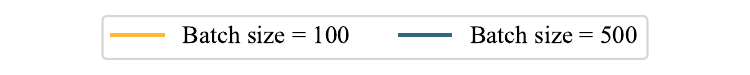}
    \end{minipage}
    \caption{Tracking performance of mini-batch SGD relative to the full-data CoLiDE-EV algorithm. The left plot illustrates the tracking of the output graph $\bbW^{\star}$, while the right plot depicts the tracking of the noise level $\sigma^{\star}${~\cite{saboksayr2023colide}}.}
    \label{fig:App_batch}
        \vspace{-10pt}
\end{figure}

\noindent {\bf Online updates.} 
{The separable CoLiDE score function lends itself to mini-batch or online optimization, because the updates decompose in time (i.e., across samples). To fix ideas, consider a homoscedastic mini-batch setting. At each iteration $t$ suppose we have access to a randomly sampled subset of data $\bbX_t \in \reals^{d \times n_b}$, with $n_b < n$ denoting the size of the mini-batch. Naturally, one could bring to bear stochastic gradient descent (SGD) iterations within the inexact BCD framework. Given an initial $\hat{\mathbf{W}}_{0}$ and only storing \emph{sufficient statistics} for the noise estimate, 
the online updates would proceed as follows. We first compute the residual $\epsilon_t = \frac{1}{n_b d} \| \bbX_t - \hat{\mathbf{W}}^{\top}_{t-1} \bbX_t \|_{F}^{2}$ and update the sufficient statistic $e_t = e_{t-1} + \epsilon_{t}$. Finally, we form $\smash{\hat{\sigma}_t = \max\left( \sqrt{e_t/t}, \sigma_0 \right)}$ which is used to update $\hat{\mathbf{W}}_{t}$ via a first-order SGD step as in Algorithm~\ref{A:colide}. Such recursive sufficient statistics have a long-standing tradition in the adaptive SP and online learning literature~\cite{mairal2010online}. Beyond this conceptual sketch, several important details are yet to be addressed to achieve a fully online solution.} Promising preliminary experimental results {reported in \cite{saboksayr2023colide}, and reproduced here in} Figure~\ref{fig:App_batch}, demonstrate that mini-batch SGD with varying batch sizes adeptly follows the output of CoLiDE-EV for both the DAG adjacency matrix (left) and the noise level (right).


\begin{table}[t]
\caption{DAG recovery performance on the Sachs dataset \cite{sachs}. The ground-truth DAG is shown in the right panel.}\label{tab:real}
\begin{minipage}[c]{0.74\textwidth}
\makebox[\textwidth][c]{
    \begin{tabular}{l @{\hspace{0em}} ccccccc}
\hline
    & {\scriptsize GOLEM-NV} & {\scriptsize DAGMA} & {\scriptsize SortNRegress}  & {\scriptsize GES} & {\scriptsize CoLiDE-EV} & {\scriptsize CoLiDE-NV} & {\scriptsize NOMAD}\\ \hline\hline
{\footnotesize SHD} & {\footnotesize 15} & {\footnotesize 16}   & {\footnotesize 13}    & {\footnotesize 13} & {\footnotesize 13}   & {\footnotesize \underline{12}} & {\footnotesize\bf 10} \\
{\footnotesize SID} & {\footnotesize 58} & {\footnotesize 52}   & {\footnotesize 47}    & {\footnotesize 56} & {\footnotesize 47}   & {\footnotesize\bf 46} & {\footnotesize -} \\
{\footnotesize FDR}   & {\footnotesize 0.66}  & {\footnotesize \underline{0.5}} & {\footnotesize 0.61}  & {\footnotesize \underline{0.5}} & {\footnotesize 0.54} & {\footnotesize 0.53} & {\footnotesize \textbf{0}}\\
{\footnotesize TPR}   & {\footnotesize 0.11}  & {\footnotesize 0.05} & {\footnotesize 0.29}  & {\footnotesize 0.23} & {\footnotesize 0.29} & {\footnotesize \underline{0.35}} & {\footnotesize \textbf{0.41}}\\
\hline
\end{tabular}
}
\end{minipage}
\begin{minipage}[c]{0.25\textwidth}
    \includegraphics[width=\textwidth]{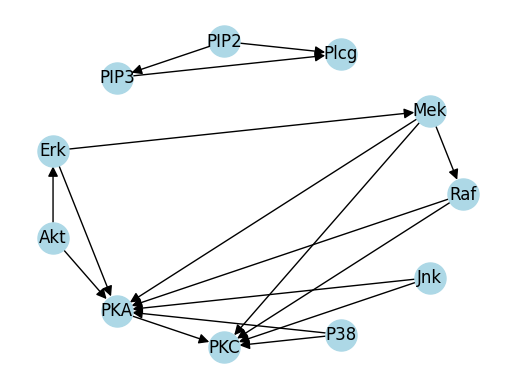}
    \end{minipage}
\vspace{-10pt}
\end{table}%

\noindent \textbf{Cell-signaling data.} We conclude our numerical analysis with a real-world application involving cell-signaling data. We study the Sachs dataset~\cite{sachs}, which is used extensively throughout the probabilistic graphical models' literature and collects cytometric measurements of protein and phospholipid constituents in human immune system cells~\cite{sachs}. It comprises $d=11$ nodes and $n=853$ samples{, and data are standardized as suggested in~\cite{sortnregress}}. The associated DAG (shown in the right panel of Table~\ref{tab:real}) was experimentally determined and validated by the biological research community; details can be found in~\cite{sachs}. The ground-truth DAG $\ccalG$ consists of $|\ccalE|=17$ edges. The outcomes of this experiment are consolidated in Table~\ref{tab:real}. The results illustrate that CoLiDE-NV outperforms all other state-of-the-art methods we have encountered so far, achieving a SHD of $12$. 

The best overall performance is attained by NOMAD~\cite{rey2026nonnegativity}, an algorithm we will describe in the following section. Unlike all other baselines, NOMAD leverages the additional constraint on edge weights being non-negative -- a characteristic of the Sachs DAG and several other settings in causal discovery. 
{In all fairness, it should be said that all these methods end up being fairly conservative when it comes to edge detection, as manifested by the abysmally low TPR values reported. This structure identification problem is known to be quite challenging and results should be interpreted with care in light of concerns (e.g., the ground-truth neglects signaling feedback) raised with the Sachs benchmark~\cite{mooij2013uai}. For other valuable real-world datasets that can be used for evaluation; see e.g.,~\cite{B_N_R}. }




\section*{Non-negative edge weights}
\label{sec:nonnegative}

Let us re-examine the role of the acyclicity function $\ccalH(\bbW)$. As discussed in the ``Continuous optimization'' section, most smooth acyclicity characterizations rely on the product $\bbW \circ \bbW$, which has two important shortcomings.
First, the nonconvex Hadamard product complicates the optimization landscape. Second, the corresponding gradients vanish at every DAG, i.e., $\nabla \ccalH(\bbW) = \bbzero$ for every adjacency matrix $\bbW$ such that $\ccalG(\bbW)\in\mathbb{D}$.
This yields degenerate Karush-Kuhn-Tucker (KKT) conditions: for a DAG $\bbW$ to be a valid KKT point, it must also be an unconstrained stationary point of the score function~\cite{noFears}. 

To overcome these limitations, a simple yet effective alternative is to specialize DAG structure learning to non-negative edge weights~\cite{rey2026nonnegativity}.
This assumption is appropriate when directed influences are known to be sign-definite, as in excitatory neural or economic influence networks, certain flow, pollution, or contagion models, and in systems where inhibitory effects are absent or can be separated beforehand. {Feasibility of such `rectifications' must be informed by domain knowledge. As with any model, deviation from postulated laws (say, if inhibitory effects are missed) will unavoidably lead to performance degradation. In other contexts beyond causal discovery, imposing non-negativity can be a convenient \emph{design choice} for subsequent construction of meaningful orthogonal Fourier bases~\cite{mihal2026icassp}.}
Inspired by DAGMA's log-determinant penalty in \eqref{eq:dagma}, non-negativity enables an exact characterization of acyclicity over the domain $\mbW_+^s = \{ \bbW \in \reals_+^{d \times d}: s > \rho(\bbW) \}$ through the simplified function 
\begin{equation}\label{eq:nonneg_logdet_constraint}
    \ccalH_+(\bbW; s) = d\log(s)-\log\det(s\bbI-\bbW).
\end{equation}
%
The intuition follows from the series expansion
\begin{equation}\label{eq:ldet_trace_series}
    \ccalH_+(\bbW; s) = -\tr \left( \log \left( \bbI - s^{-1}\bbW \right) \right) =
    \sum_{k=1}^{\infty}\frac{\tr(\bbW^k)}{k s^k},
\end{equation}
where non-negativity prevents sign cancellations in $\tr(\bbW^k)$.
Consequently, $\ccalH_+(\bbW; s) = 0$ if and only if $\ccalG(\bbW)\in\mathbb{D}$ for $\bbW\in\mbW_+^s$.
Identity~\eqref{eq:ldet_trace_series} also sheds light on the role of the parameter $s$ in DAGMA-like constraints, with smaller values of $s$ penalizing longer directed paths more strongly. Indeed, for $s=1$ notice how the slower $1/k$ modulation in \eqref{eq:ldet_trace_series} offers better discriminability to detect longer cycles than, say, the NOTEARS function $\ccalH_{\text{expm}}(\bbW)$ which decays like $1/k!$. Going back to the optimality conditions, DAGs are no longer stationary points of $\ccalH_+(\bbW; s)$, avoiding the aforementioned KKT degeneracy.

Using the prototypical OLS score for linear SEMs in~\eqref{eq:ols_l1_loss} together with the non-negativity constraints and the acyclicity characterization $\ccalH_+(\bbW; s)$, the DAG learning problem can be cast as
\begin{equation}\label{eq:nonneg_dag_learning}
    \hbW = \operatorname*{arg\,min}_{\bbW}
    \underbrace{\frac{1}{2n}\| \bbX-\bbW^\top\bbX \|_F^2 + \lambda \sum_{i, j} W_{ij}}_{:=\ccalS_+(\bbW;\bbX)}
    \;\;
    \text{subject to } \;
    \bbW \ge 0, \; \rho(\bbW) < s, \;
    \ccalH_+(\bbW; s) {=} 0,
\end{equation}
where the $\ell_1$-norm reduces to a linear penalty since $W_{ij}\geq 0$. 
Although the problem remains nonconvex, the combination of non-negativity and the acyclicity function $\ccalH_+(\bbW; s)$ leads to a more benign optimization landscape\cite{rey2026nonnegativity}.
Moreover, the added structure justifies the adoption of principled algorithms for constrained optimization, the subject we deal with next.

\subsection*{Algorithmic implementation and landscape analysis}
\label{ssec:nonneg_optimization}
The absence of degenerate KKT conditions in \eqref{eq:nonneg_dag_learning} motivates the use of the MoM, an augmented Lagrangian-based iterative solver tailored to equality-constrained optimization problems.
Denote the objective function in \eqref{eq:nonneg_dag_learning} by $\ccalS_+(\bbW; \bbX)$.
For a multiplier $\alpha \in \reals$ and a penalty parameter $c > 0$, the augmented Lagrangian of \eqref{eq:nonneg_dag_learning} is given by
\begin{equation}\label{eq:augmented_lagrangian}
    \ccalL_c(\bbW, \alpha) = \ccalS_+(\bbW; \bbX) + \alpha \ccalH_+(\bbW; s) + \frac{c}{2} \ccalH_+^2(\bbW; s),
\end{equation}
%
Notice that only the acyclicity constraint has been dualized in \eqref{eq:augmented_lagrangian}.
The non-negativity constraint is kept explicit because it can be handled directly through a projection onto the non-negative orthant. Similarly, the spectral-radius condition is treated as a domain restriction for the log-determinant term.


Given the augmented Lagrangian, one can employ the MoM to estimate the DAG structure.
The resulting iterative algorithm is termed Non-negative Optimization via Multipliers for Acyclic Digraphs (NOMAD), which we encountered with the Sachs dataset.
Starting from $\alpha_0$ and $c_0>0$, NOMAD iterates over the following three steps for at most $k=1,\ldots,T_{\text{max}}$ iterations or until a stopping criterion is met.
First, it updates the DAG estimate $\bbW_{k+1}$ by minimizing the augmented Lagrangian $\ccalL_{c_{k}}(\bbW,\alpha_{k})$, which can be solved with inner projected first-order iterations.
Second, it updates the multiplier $\alpha_{k+1}$ using a dual gradient-ascent step, since the gradient w.r.t. $\alpha$ is precisely the constraint violation residual $\ccalH_+(\bbW_{k+1}; s)$.
Third, it increases the penalty parameter $c_{k+1}$ only when the acyclicity residual does not decrease sufficiently. The resulting MoM procedure follows the rationale of classical penalty methods: the quadratic term penalizes violations of $\ccalH_+(\bbW; s)=0$, while the multiplier update corrects the linearized constraint across iterations. The computational cost is dominated by the inner problem in the first step, whose complexity is $\ccalO(d^3)$ just like most other continuous DAG-learning approaches.

\begin{figure}[t]
    \centering
    \begin{minipage}[c]{.32\textwidth}
    \includegraphics[width=\textwidth]{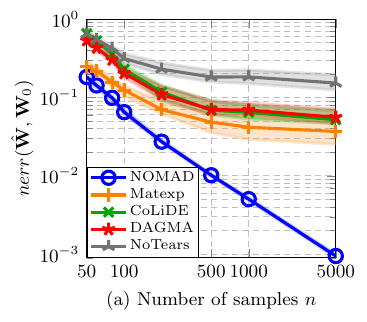}
    \end{minipage}
    \begin{minipage}[c]{.32\textwidth}
    \includegraphics[width=\textwidth]{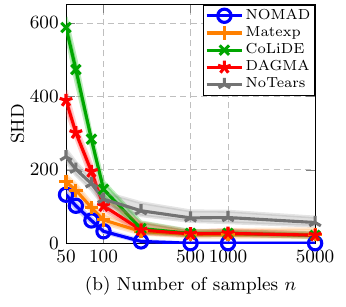}
    \end{minipage}
    \begin{minipage}[c]{.32\textwidth}
    \includegraphics[width=\textwidth]{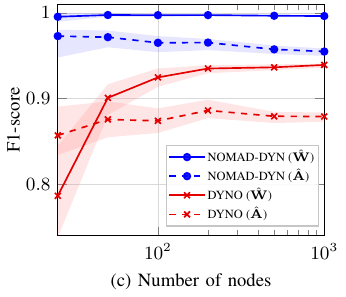}
    \end{minipage}
    \caption{Recovery of ER4 DAGs with non-negative weights: (a)--(b) use $d=100$, while (c) uses $t_{\text{max}}=5000$, $\tau_{\text{max}}=2$, and $\bbA_\tau$ with mean degree $2$. Curves report median metrics, with $25$th--$75$th percentiles, over $50$ realizations. {Adapted from~\cite{rey2026nonnegativity}.}}
    \label{fig:exps_nomad}
    \vspace{-10pt}
\end{figure}

Although nonconvexity precludes a global recovery guarantee, empirical results suggest that NOMAD recovers the true DAG when the number of samples $n$ is large enough.
This is illustrated in Figure~\ref{fig:exps_nomad}(a), where the normalized mean-squared error of the estimated $\hbW$ vanishes as $n$ increases. Similarly, Figure~\ref{fig:exps_nomad}(b) {shows that NOMAD achieves accurate support recovery,} outperforming other baselines and highlighting the benefits of exploiting this additional structure; recall the Sachs results in Table~\ref{tab:real}. 

To contextualize these empirical findings, we comment on the optimization landscape of the population augmented Lagrangian, where the empirical OLS loss is replaced by its expectation. Interestingly, non-negativity induces a benign landscape with three useful properties: (i) the true DAG $\bbW_0$ is the unique global minimizer of the augmented Lagrangian; (ii) there are no spurious interior stationary points; and (iii) every acyclic KKT point is $\bbW_0$; see~\cite[Sec. IV]{rey2026nonnegativity} for details. Although these properties do not guarantee convergence of NOMAD to the global optimum or the true DAG, they describe a favorable scenario and illustrate the value of harnessing additional structure when imposing acyclicity.

\subsection*{Non-negative DAG learning from time-series}\label{ssec:svarm}
The benefits and strong performance of NOMAD stem from its simpler acyclicity constraint, which hinges on exploiting the non-negativity of the DAG. Just like CoLiDE, a key benefit of this approach is its flexibility. Non-negativity can be incorporated into different acyclicity characterizations and combined with a host of score functions. To elucidate this property, here we move beyond the linear SEM in \eqref{eq:linear_sem_nodal} and consider a non-i.i.d. setting whereby the observed data form a time series described by a SVARM.

Consider the problem of inferring causal structure from a multivariate time series $\{\bbx_t\}_{t=1}^{\infty}$, where $\bbx_t \in \reals^d$ collects the measurements at discrete time instant $t$. We assume that the data-generating process is given by a SVARM of order $\tau_{\text{max}}$; see e.g.,~\cite{elements,giannakis18,mateos2019connecting}.
Formally, the signal $\bbx_t$ is modeled as
\begin{equation} \label{eq:svarm_vector}
	\bbx_t = \bbW^\top \bbx_t + \sum_{\tau=1}^{\tau_{\text{max}}} \bbA_\tau^\top \bbx_{t-\tau} + \bbz_t,
\end{equation}
where $\bbW$ encodes instantaneous dependencies, and the $d\times d$ matrices $\{\bbA_\tau\}_{\tau=1}^{\tau_{\text{max}}}$ capture time-lagged influences from the $\tau$-th previous step.
Again, $\bbz_t$ is a vector of mutually independent exogenous noises.

Under model \eqref{eq:svarm_vector}, inference amounts to estimating the instantaneous matrix $\bbW$ and the lagged matrices $\bbA_\tau$. For a valid causal interpretation in the absence of latent variables, $\bbW$ must be a DAG.
On the other hand, the lagged coefficient matrices $\bbA_\tau$ need not be acyclic, as they describe how past system states influence the current one at time $t$. Once an appropriate score function is chosen, such as sparsity-penalized OLS in \eqref{eq:ols_l1_loss} or CoLiDE in \eqref{eq:genral-colide-ev}, acyclicity of $\bbW$ can be imposed through any of the functions discussed in the ``Continuous optimization'' section. Here again, if negative or repressive relations can be ruled out, enforcing non-negativity will improve causal discovery performance. Figure~\ref{fig:exps_nomad}(c) illustrates the dividends from exploiting non-negativity in the SVARM setting, where we observe a near-perfect F1-score in recovery of $\bbW$ (and improved estimation of $\bbA_\tau$) relative to a non-negativity agnostic baseline.




\section*{Discussion and the road ahead}
\label{sec:conclusions}

DAGs constitute a central graphical modeling tool at the heart of SEMs and related Bayesian networks, with well-documented merits to enable principled reasoning about cause-effect interactions that govern complex systems. Causal understanding is central to almost every endeavor of scientific inquiry~\cite{Pearl_2009,elements}. It is also a burgeoning theme in the development of ML and AI technologies, as (correlation-based) statistical models trained on i.i.d. data are insufficient to address open problems related to transfer learning and (out-of-domain) generalization~\cite{crl2021pieee}. Since the causal structure underlying a group of variables is often unknown and interventions may be infeasible to implement due to resource constraints or ethical considerations, there is a need to address the NP-hard task of inferring DAGs from observational data. However, most classical structure identification approaches face two key obstacles: the combinatorial challenge of enforcing acyclicity, which severely limits scalability, and identifiability challenges arising from latent confounding, finite sample effects, or heterogeneous noise profiles in workhorse SEMs. 

In this context, this article offered an overview of recent SP and optimization advances to address these issues by recasting DAG structure learning as a continuous, score-based estimation problem over adjacency matrices. We began with a didactic introduction to SEMs and the formulation of causal graph discovery, followed by a historical survey of score-based methods ranging from early combinatorial search schemes and greedy heuristics to modern continuous frameworks that leverage smooth characterizations of acyclicity. {Building on this foundation, we emphasized concomitant DAG estimators that jointly infer sparse causal structure and exogenous noise levels, with the goal of improving robustness under heteroscedasticity and distribution shifts by rendering the estimator noise adaptive. Within the broader and vibrant landscape of score-based continuous optimization methods, CoLiDE offers an alternative to vanilla lasso-type score functions which is in practice most beneficial whenever the unknown (and possibly unequal) exogenous noise scales change across problem instances. In these settings, beyond experimental observations on improved estimation of edge weights, the concomitant score requires minimal to no sparsity-regularization parameter retuning. Unlike GOLEM, noise adaptivity does not sacrifice problem decomposability across samples (crucial for scalable stochastic mini-batch or online algorithms), and it is not tied to Gaussian distributional assumptions.} When pertinent, we argued that exploiting non-negative structure in the edge weights can pay optimization, computational, and statistical dividends. {Going back to the continuous formulation in \eqref{eq:discrete_cont}, CoLiDE's focus is on enhancing the score $\ccalS$ while non-negativity can bring around important simplifications to the acyclicity function $\ccalH$.  The potential benefits of combining both ideas are worthy of further exploration~\cite{CoLiDE_ICASSP_27}.}

Admittedly, the scope of the surveyed CoLiDE framework has been limited to observational data adhering to a linear SEM. Extensions to encompass nonlinear and (soft) interventional settings~\cite{xue2023dotears} are certainly of interest. Therein, CoLiDE's formulation, amenable to first-order optimization, will facilitate a symbiosis with neural networks or kernel methods to parameterize SEM nonlinearities; see e.g.,~\cite{giannakis18}. We have also implicitly relied on the strong assumption of sufficiency, i.e., there were no latent variables, and thus all relevant data could be observed. In the presence of hidden confounding, causal statements cannot be inferred from estimated DAGs, and a latent variable modeling substrate may be called for.

\subsection*{Research outlook}
\label{ssec:research_outlook}

A wide variety of potential research avenues naturally
follows from the developments presented here. Although CoLiDE's decomposability (across variables and samples) is a demonstrable property, further work is needed to fully assert the practical value of stochastic mini-batch optimization in large-scale settings. In terms of computational complexity, there is room for
improving the scalability of some of the solvers described via
parallelization and decentralized implementations. Moreover,
online adaptive algorithms that can track the (possibly) time-varying connectivity structure of the acyclic network and achieve both memory and computational savings by processing the signals on-the-fly are naturally desirable, but so far largely unexplored. In terms of optimization theory and algorithms, we envision multiple avenues to realize CoLiDE's full potential both in batch and online settings, with anticipated impact also to order-based methods that search in the space of vector orderings~\cite{topo} or matrix permutations~\cite{permutahedron}. Although global optimality results are so far elusive for the methods in the ``Continuous optimization'' section, there is hope as progress on benign optimization landscape analyses is being made~\cite{deng2023global,rey2026nonnegativity}. 

The surveyed approaches enforce acyclicity via differentiable penalties but do not incorporate learned priors over graph structure -- an untapped opportunity worthy of exploration. Cross-pollinating advances from computational imaging, a plug-and-play algorithm was put forth in~\cite{DAGPnP_Asilomar_26} that combines gradient steps on a linear SEM data-fidelity objective with a DAGMA penalty, and then applies denoising steps from a diffusion model trained on weighted DAGs to bias the solution towards plausible digraph topologies.

In many applications, the recovered DAG is not an end in itself, but is passed to a downstream learner (for prediction, simulation, or control~\cite{rey2026dcn}), in which case structural-recovery errors can propagate into and degrade task performance. This two-step sequential learning pipeline (structure first, task second) is arguably the rule rather than the exception in graph-aware ML, either by convention; or, due to the convenience of modular separation between structure-learning and task-execution components. In this context, it is also of interest to develop \emph{task-aware} DAG learning frameworks that incorporate the downstream task loss directly into the structure-identification objective, jointly optimizing for graph fidelity and task utility (or, e.g., fairness metrics) subject to acyclicity constraints. The envisioned problem is naturally a multi-objective optimization; hence, amenable to scalarization or multiple-gradient descent algorithms for which descent directions can be analytically derived and used to update graph and task parameters in tandem. Bilevel optimization advances can also play a key role; see, e.g., the general causal learning model inspired by meta-learning proposed in~\cite{metadag2023icassp}, which aims at finding an invariant DAG over multiple domains and increasing the generalization performance of DAG structure discovery.

%
\section*{Acknowledgment}
This work was supported by the NSF under Award ECCS 2231036, by the Spanish AEI Grants PID2022-136887NB-I00 and PID2023-149457OB-I00, and by the Community of Madrid (via grants CAM-URJC F1180 (CP2301), TEC-2024/COM-89, and Madrid ELLIS Unit).

%
\bibliographystyle{IEEEtranS}
\bibliography{refs}

%
%
%
%
%
\section*{Biographies}
\label{sec:bio}
\vspace{-2cm}
\begin{IEEEbiographynophoto}{Gonzalo Mateos}
received his B.Sc. degree in Electrical Engineering from Universidad de la Rep\'ublica, Montevideo, Uruguay in 2005 and the M.Sc. and Ph.D. degrees in Electrical Engineering from the University of Minnesota, Minneapolis, in 2009 and 2012. Currently, he is a Professor with the Department of Electrical and Computer Engineering, University of Rochester, as well as the Associate Director for Research at the Goergen Institute for Data Science and Artificial Intelligence. He also was an Asaro Biggar Family Fellow in Data Science (2020-23). His research interests lie in the areas of statistical learning from complex data, network science, decentralized optimization, and graph signal processing, with applications in brain connectivity, causal discovery, wireless network monitoring, power grid analytics, and information diffusion. He is a Senior Member of the IEEE.
\end{IEEEbiographynophoto}
\vspace{-2cm}

\begin{IEEEbiographynophoto}{Samuel Rey} received the B.Sc., M.Sc., and Ph.D. degrees in telecommunication engineering from King Juan Carlos University (URJC), Madrid, Spain, in 2016, 2018, and 2023, respectively, all with highest honors.
In 2023, he joined the Department of Signal Theory and Communications, URJC, where he is currently an Assistant Professor. His research interests include graph signal processing, machine learning, non-convex optimization, and data science over networks. Dr. Rey has served on the organizing committees of several international conferences. He received the Best Young Investigator Award from URJC in 2018 and was awarded the Spanish National FPU Scholarship for Ph.D. studies in the same year. He is a Member of the IEEE.
\end{IEEEbiographynophoto}
\vspace{-2cm} 

\begin{IEEEbiographynophoto}{Hamed Ajorlou}
received the B.Sc. degree in Electrical Engineering from Sharif University of Technology, Tehran, Iran, in 2023 and the M.Sc. degree in Electrical Engineering from the University of Rochester, Rochester, NY, in 2025. Since 2023, he has been working toward the Ph.D. degree at the University of Rochester. His research interests are in developing efficient algorithms for large-scale graph processing and exploring the theoretical foundations of graph neural networks. He is a Graduate Student Member of the IEEE.
\end{IEEEbiographynophoto}
\vspace{-2cm} 

\begin{IEEEbiographynophoto}{Mariano Tepper} received his B.Sc. and Ph.D. degrees in Computer Science from the Universidad de Buenos Aires, Argentina in 2006 and 2011. In between, he received the M.Sc. degree in Applied Mathematics from the \'Ecole Normale Sup\'erieure de Cachan, France, in 2007. Currently, he is a Senior Principal Engineer at Elastic. His main research interest is semantic search, developing indices and representations for large-scale vector search, while also delving into unsupervised machine learning and causal discovery.
\end{IEEEbiographynophoto}

\vfill

\end{document}